\documentclass[conference]{IEEEtran}
\IEEEoverridecommandlockouts
\usepackage{cite}
\usepackage{amsmath,amssymb,amsfonts}
\usepackage{algorithmicx}
\usepackage{graphicx}
\usepackage{textcomp}
\usepackage{xcolor}

\usepackage{microtype}

\usepackage{inconsolata}
\usepackage[ruled,lined,linesnumbered]{algorithm2e}
\usepackage{csquotes}

\usepackage{microtype}
\usepackage{inconsolata}
\usepackage{bm}
\usepackage{amsfonts}
\usepackage{todonotes}
\usepackage[numbers]{natbib}
\usepackage{graphicx}
\usepackage{footnote}
\usepackage{color}
\usepackage{amsfonts}
\usepackage{multicol}
\usepackage{multirow}
\usepackage{amssymb}
 \usepackage{enumitem}
\usepackage{color, colortbl}
\usepackage{algpseudocode}
\usepackage{soul}
\usepackage{arydshln}
\usepackage{balance}
\usepackage{flushend}
\usepackage{tabularx,booktabs}

\usepackage{caption}
\usepackage{subcaption}

\usepackage[hidelinks]{hyperref}

\usepackage{array}
\newcolumntype{P}[1]{>{\centering\arraybackslash}p{#1}}
\newcolumntype{M}[1]{>{\centering\arraybackslash}m{#1}}

\def\BibTeX{{\rm B\kern-.05em{\sc i\kern-.025em b}\kern-.08em
    T\kern-.1667em\lower.7ex\hbox{E}\kern-.125emX}}
\begin{document}


\title{\textsc{GPT-Detox}: An In-Context Learning-Based \\ Paraphraser for Text Detoxification}

\author{\IEEEauthorblockN{Ali Pesaranghader$^*$}
\IEEEauthorblockA{\textit{LG Electronics, Toronto AI Lab} \\
Toronto, Canada \\
ali.pesaranghader@lge.com}
\and
\IEEEauthorblockN{Nikhil Verma$^*$} 
\IEEEauthorblockA{\textit{LG Electronics, Toronto AI Lab} \\
Toronto, Canada \\
nikhil.verma@lge.com}
\and
\IEEEauthorblockN{Manasa Bharadwaj$^*$}
\IEEEauthorblockA{\textit{LG Electronics, Toronto AI Lab} \\
Toronto, Canada \\
manasa.bharadwaj@lge.com}
}

\maketitle
\def\thefootnote{*}\footnotetext{Equal Contribution}\def\thefootnote{\arabic{footnote}}

\begin{abstract}

Harmful and offensive communication or content is detrimental to social bonding and the mental state of users on social media platforms. 
Text detoxification is a crucial task in natural language processing (NLP), where the goal is removing profanity and toxicity from text while preserving its content. 
Supervised and unsupervised learning are common approaches for designing text detoxification solutions. However, these methods necessitate fine-tuning, leading to computational overhead. 
In this paper, we propose \textsc{GPT-Detox} as a framework for prompt-based in-context learning for text detoxification using GPT-3.5 Turbo. We utilize zero-shot and few-shot prompting techniques for detoxifying input sentences. 
To generate few-shot prompts, we propose two methods: word-matching example selection (WMES) and context-matching example selection (CMES). 
We additionally take into account ensemble in-context learning (EICL) where the ensemble is shaped by base prompts from zero-shot and all few-shot settings.
We use ParaDetox and APPDIA as benchmark detoxification datasets.
Our experimental results show that the zero-shot solution achieves promising performance,
while our best few-shot setting outperforms the state-of-the-art models on ParaDetox and shows comparable results on APPDIA.
Our EICL solutions obtain the greatest performance, adding at least 10\% improvement, against both datasets.
%

\end{abstract}

\section{Introduction}

\textit{\textbf{Disclaimer.} You may encounter toxic text while reading this paper due to the nature of the work.}

Harmful and offensive communication or content is detrimental to social bonding and the mental state of the users on social media platforms \cite{gonzalez2023social}. Therefore, text detoxification, with the goal of removing profanity and toxicity from text while preserving its content, has become an important task in natural language processing (NLP). 

Large language models (LLMs) have become the center of attention due to their promising results in various NLP applications \cite{zhao2023survey}. Despite that, LLMs may generate toxic and harmful outputs because they are trained to predict the next word on a large dataset of the Internet without being instructed \cite{weidinger2021ethical}.
Besides, conventional fine-tuning, for a specific task, may not reduce their toxicity \cite{brown2020language}.
To mitigate the issue above, we need to instruct LLMs when they are trained.
As a recent effort, InstructGPT \cite{ouyang2022training} (also known as GPT-3.5) relies on reinforcement learning from human feedback (RLHF) \cite{christiano2017deep} technique to generate safer and more aligned outputs to user's intentions. The RLHF technique uses human preferences as a reward signal to fine-tune the models, which is important for safety and alignment that are complex, subjective, and not fully captured by simple automatic metrics.

In the literature, supervised and unsupervised solutions are proposed for text detoxification \cite{li2018delete, wu2019mask, lee2020stable, dale2021text}. Those solutions, however, may introduce a dependency on a detoxification model, leading to more computational overhead.
Compared to supervised and unsupervised solutions, in-context learning (ICL) via LLMs leads to a significant reduction in the amount of task-specific data. This reduction helps mitigate overfitting to fine-tuning datasets. Furthermore, in-context learning eliminates the need for additional computational resources dedicated to fine-tuning \cite{brown2020language}. 

To the best of our knowledge, in-context learning using instruction-tuned LLMs has  not been used for text detoxification in the literature. This motivated us to consider prompt-based in-context learning for text detoxification in this study. 

%

In this work, we introduce \textsc{GPT-Detox} as a prompt-based in-context learning framework for text detoxification which uses GPT-3.5\footnote{\href{https://platform.openai.com/docs/models/gpt-3-5}{GPT-3.5 Turbo}}\textsuperscript{,}\footnote{We had access to only GPT-3.5 Turbo at the time of running our experiments. GPT-4 may be used in future research work.} as a generative model.
We study the effectiveness of GPT-3.5 for detoxification, 
against two benchmark parallel datasets, namely ParaDetox \cite{logacheva2022paradetox} and APPDIA \cite{atwell2022appdia}. 
Note that no direct mapping exists between sentences of the source and target style in non-parallel datasets; therefore, it is not possible to consider them for few-shot style-transfer prompting where we need a few mapping examples. 
Also, in the case of non-parallel data, models only receive a weak supervision of the target sentences which may result in style-transfer failures \cite{liu2022nonparallel}. For those reasons, we consider only the parallel datasets in this study.


Our experiments show that our detoxified sentences are more accurate, similar, and fluent compared to those generated by the state-of-the-art models. We summarize our contributions as follows:

\begin{itemize}[noitemsep]
    \item We introduce \textsc{GPT-Detox} as a framework for zero-shot and few-shot prompting as well as ensemble in-context learning (EICL) for text detoxification.
    \item We define word-matching example selection (WMES) and context-matching example selection (CMES) as two approaches for example selection for few-shot prompting.
    \item We present ensemble prompting where the ensemble is shaped by base prompts from zero-shot and all few-shot settings.   
    \item We conduct an in-depth analysis of the detoxified sentences with respect to different evaluation metrics.
    \item We show that \textsc{GPT-Detox} outperforms the state-of-the-art models against the ParaDetox and APPDIA datasets.
\end{itemize}

The remainder of this paper is as follows. We discuss related work 
in Section \ref{sec:related-work}. We describe our text detoxification methodology in Section \ref{sec:methodology}. In Section \ref{sec:experiments}, we run our experiments and provide a comprehensive evaluation. Finally, we conclude our paper in Section \ref{sec:conclusion}.

\section{Related Work}
\label{sec:related-work}

\subsection{Text Detoxification}

Text detoxification is often defined as a style transfer from toxic to neutral while the content of the original input is preserved \cite{logacheva2022paradetox}. 
%
The detoxification solutions usually fall into two main categories, \textit{unsupervised} and \textit{supervised}.
The unsupervised methods are built on non-parallel datasets. 
%
Mask\&Infill \cite{wu2019mask}, DRG-Template/Retrieve \cite{li2018delete}, DLSM \cite{he2020probabilistic}, SST \cite{lee2020stable}, CondBERT and ParaGeDi \cite{dale2021text} fall into this category.
On the other hand, the supervised methods are trained on parallel datasets in which one-to-one mappings between toxic and non-toxic texts exist. For text detoxification, 
\citet{logacheva2022paradetox} fine-tuned BART \cite{lewis2020bart} on their parallel data, i.e., ParaDetox, and showed that their model outperforms the unsupervised methods. 
Recently, \citet{floto2023diffudetox} introduced DiffuDetox, a mixed conditional and unconditional diffusion model for text detoxification. Their experiments showed that their model achieves a comparable performance to the ParaDetox model and other unsupervised models.

\subsection{Prompt-based In-Context Learning}

\citet{brown2020language} introduced prompt-based in-context learning, also known as few-shot prompting, as a new learning paradigm where LLMs, e.g., GPT-3 variants, are prompted to generate a (textual) output. 
This paradigm has shown promising results in various NLP tasks, even outperforming state-of-the-art fine-tuning approaches \cite{brown2020language, schick2021its}.
\citet{liu-etal-2022-makes} showed that selecting semantically similar examples for in-context learning improves responses by GPT-3 for open-domain question-answering (QA) tasks.
\citet{wei2021finetuned} achieved good performance on unseen tasks by using instruction tuning with LaMBDA-PT.
\citet{meng2022generating} and \citet{yoo2021gpt3mix} obtained promising performance for different classification tasks by fine-tuning language models against task-specific datasets created using informative prompts. 
\citet{zheng2022exploring} and \citet{madotto2021fewshot} successfully applied few-shot prompting in dialogue generation tasks.
\citet{yeinvestigating} explores ensembling in-context learning for the task of question answering by leveraging late fusion.
The main advantage of in-context learning is that no training or fine-tuning is required when compared to unsupervised and supervised solutions.
However, engineering prompts and in-context examples have a direct impact on performance.
To approach this problem, \citet{liu-etal-2022-makes} proposed KATE, a KNN-based approach, for in-context example selection. Later, \citet{lee2022does} extended KATE and introduced \textsc{SitSM} and \textsc{EmoSitSM} to generate empathetic responses using GPT-3 after selecting relevant examples using emotion and situation.

\section{Methodology}
\label{sec:methodology}
We propose zero-shot and few-shot prompting for text detoxification using GPT-3.5. We describe our methodology for each, in the following subsections.

\subsection{Zero-Shot Prompting}

Zero-shot prompting refers to providing a prompt, that does not contain any example from the training data, to guide the generation of desired output. This approach has made LLMs useful for many NLP tasks \cite{lee2022does, ye2023context, shen2023flan, wang2023plan}. Hence, we consider zero-shot prompting as our first approach for text detoxification. Our zero-shot prompt template and a corresponding example are shown in Table \ref{tab:zero-shot-prompting-methodology}.

\begin{table*}[t]
    \centering
    \resizebox{\linewidth}{!}{
        \begin{tabular}{l|p{12cm}}
        \hline
        Template & Rephrase the following sentence by only replacing offensive, harmful, and swear words while using the remaining words: [INPUT SENTENCE] \\ \hline
        Prompt & Rephrase the following sentence by only replacing offensive, harmful, and swear words while using the remaining words: \textcolor{red}{\hl{this sick fuck is just a sociopath} who doesn't want to face the consequences for his actions.} \\ \hline
        \end{tabular}%
    }
    \caption{Zero-Shot Prompting. The red sentence is the toxic input, and the highlight shows the toxic part.}
    \label{tab:zero-shot-prompting-methodology}
\end{table*}

\subsection{Few-Shot Prompting}

Few-shot prompting enables in-context learning where we provide a few examples in the prompt to help LLMs understand the task better for generating suitable outputs when compared to zero-shot prompting.
In this section, we introduce example selection methods for few-shot prompting: (1) word-matching example selection (WMES), and (2) context-matching example selection (CMES). 
In addition to WMES and CMES, we also have random few-shot prompting where examples are selected randomly from the training pairs of input and reference sentences.
We use $k$ to denote the number of examples in a prompt, and we consider $k \in \{1, 3, 5, 8, 10\}$ in this study.
%
Table \ref{tab:few-shot-prompting-methodology} shows our few-shot prompt template with an example.

\begin{table*}[t]
    \centering
    \resizebox{\linewidth}{!}{
    \begin{tabular}{p{1.1cm}|p{12cm}}
        \hline
        Template &  
        Paraphrase by replacing offensive, harmful, and swear words with respectful words. Follow the examples below:
        \newline \newline
        input: [INPUT SENTENCE 1] \newline
        output: [REFERENCE SENTENCE 1] 
        \newline
        ...
        \newline
        input: [INPUT SENTENCE $k$] \newline
        output: [REFERENCE SENTENCE $k$] 
        \newline \newline
        input: [TOXIC INPUT] \newline
        output:  \\ \hline
        Example ($k$ = 3) & Paraphrase by replacing offensive, harmful, and swear words with respectful words. Follow the examples below:
        \newline \newline
        input: \textcolor{red}{That's literally what you said. That's \hl{how stupid you are}.} \newline
        output: That's literally what you said. 
        \newline \newline
        input: \textcolor{red}{Congratulations on your \hl{stupid retort}. You sound special, but not in a positive way} \newline
        output: Congratulations on your pointless retort. You sound special, but not in a positive way
        \newline \newline
        input: \textcolor{red}{Your as \hl{dumb as a rock}. } \newline
        output: That is not very smart. 
        \newline \newline
        input: \textcolor{red}{You're \hl{ridiculous and stupid}. That's all you get from me.} \newline
        output: \\
        \hline 
    \end{tabular}
    }
    \caption{Few-Shot Prompting. For $k$ examples, the input and output pairs show the toxic and non-toxic sentences. The red sentences are the toxic examples, and the highlights show the toxic parts.}
    \label{tab:few-shot-prompting-methodology}
\end{table*}

\subsubsection{Word-Matching Example Selection}

The word-matching example selection (WMES) method collects the examples based on toxic word-matching between the current toxic input and the existing toxic examples from the training set. That is, $k$ training examples that share the same toxic word or phrase with the input toxic sentence are considered as candidates. Algorithm \ref{algo:word-matching} summarizes this method.

\begin{algorithm}[t]

    \caption{Word-Matching (WMES)}
    \label{algo:word-matching}
    
    \SetKwInOut{Input}{Input}
    \SetKwInOut{Output}{Output}
    
    \Input{Training data $\mathcal{D} = \{(x_i, y_i)\}_{i=1}^{N}$,\\ number of in-context examples $k$,\\ a toxic sentence $s_{toxic}$} 
    \Output{$\mathcal{L}_{k}$ $\Leftarrow$ $k$ similar examples to $s_{toxic}$} 
    
    \textcolor{gray}{\fontfamily{lmtt}\selectfont{/* Initiate a list to hold similar examples */}} \\
    $\mathcal{L}$ $\leftarrow$ \mbox{empty list} \\
    \textcolor{gray}{\fontfamily{lmtt}\selectfont{/* Retrieve similar examples */}} \\
    $\mathcal{T}$ $\leftarrow$ \mbox{\textit{toxic words} in $s_{toxic}$} \\
    \For{$(x_i, y_i) \in \mathcal{D}$}{
      \If{$x_i$ contains any word from $\mathcal{T}$}{
            $\mathcal{L}.append((x_i, y_i))$
       }
    }
    \textcolor{gray}{\fontfamily{lmtt}\selectfont{/* Output */}} \\
    $\mathcal{L}_{k}$ $\leftarrow$ Pick $k$ examples from $\mathcal{L}$ \\
    
\end{algorithm}


\subsubsection{Context-Matching Example Selection}

Context-matching example selection (CMES) relies on semantic similarity between the current toxic input and the existing toxic examples from the training set. For that, we use an MPNet-based model\footnote{Sentence Transformers: \href{https://huggingface.co/sentence-transformers/all-mpnet-base-v2}{all-mpnet-base-v2}} to encode sentences, i.e., the input and the training examples, and obtain their embedding representations. Then, we calculate the cosine similarity between the embeddings of the input and the training examples. The top-$k$ most similar examples to the input, and their non-toxic references, are used to shape our prompts (similar to the prompt in Table \ref{tab:few-shot-prompting-methodology}). Algorithm \ref{algo:context-matching} presents the pseudo-code for this method.

\begin{algorithm}[t]
    
    \caption{Context-Matching (CMES)}
    \label{algo:context-matching}
    
    \SetKwInOut{Input}{Input}
    \SetKwInOut{Output}{Output}
    
    \Input{Training data $\mathcal{D} = \{(x_i, y_i)\}_{i=1}^{N}$,\\ a sentence encoder $f_{\theta}(.)$,\\ number of in-context examples $k$,\\ a toxic sentence $s_{toxic}$}
    \Output { $\mathcal{L}_{k}$ $\Leftarrow$ Top-$k$ similar examples}

    \textcolor{gray}{\fontfamily{lmtt}\selectfont{/* Collect embeddings of inputs in $\mathcal{D}$ and input $s_{toxic}$ */}} \\    
    $\mathcal{L}_{emb} \leftarrow \mbox{empty list}$ \\
    \For{$(x_i, y_i) \in  \mathcal{D}$}{   
        $z_{x_i} \leftarrow f_{\theta}(x_i)$\\
        $\mathcal{L}_{emb}.append(z_{x_i})$
    }     
    $z_s \leftarrow f_{\theta}(s_{toxic})$ \\
    \textcolor{gray}{\fontfamily{lmtt}\selectfont{/* Collect the Cosine similarity scores using embeddings */}} \\    
    $\mathcal{L}_{sim} \leftarrow \mbox{empty list}$ \\
    \For{$z_{x_i} \in \mathcal{L}_{emb}$}{   
        $sim_i \leftarrow \frac{z_s \cdot z_{x_i}}{ ||z_s|| \cdot ||z_{x_i}|| } $ \\
        $\mathcal{L}_{sim}.append(sim_i)$
    }

    \textcolor{gray}{\fontfamily{lmtt}\selectfont{/* Output: Collect the top-$k$ similar toxic sentences in $\mathcal{D}$ */}} \\ 
    $\mathcal{I}_k$ $\leftarrow$ indices of top-$k$ scores in $\mathcal{L}_{sim}$ \\
    $\mathcal{L}_{k} \gets \mbox{empty list}$ \\
    \For{$i \in \mathcal{I}_k$}{
        $\mathcal{L}_{k}.append(\mathcal{D}(x_{i}, y_{i}))$
    }
    
\end{algorithm}

\subsubsection{Ensemble In-context Learning}

We present ensemble in-context learning (EICL) where the ensemble is shaped by base prompts from the zero-shot and few-shot settings. As for few-shot base prompts, we consider random, WMES, and CMES settings with $k \in \{1, 3, 5, 8, 10\}$. The motivation behind this setting is to choose a prompt that generates the best non-toxic paraphrase. We formulate ensemble in-context learning by the following equation:
\begin{equation}
    p^* = \arg\max_{ p \in \mathcal{P}} (f_{s}(p))
\end{equation}

\noindent where $p^*$ is the best prompt, $\mathcal{P}$ denotes the set of all prompts from all settings, and $f_{s}$ refers to a scoring function, e.g., the similarity score.






\section{Experiments}
\label{sec:experiments}

\subsection{Experimental Settings}
\label{sec:exp_settings}

\noindent \textbf{Datasets.} We run our experiments against the ParaDetox \cite{logacheva2022paradetox} and APPDIA \cite{atwell2022appdia} datasets.

\begin{itemize}[noitemsep]
    \item \textbf{ParaDetox}\footnote{\href{https://huggingface.co/datasets/s-nlp/paradetox}{ParaDetox Dataset}} 
    contains 12,610 toxic sentences from three sources: Jigsaw\footnote{\href{https://www.kaggle.com/c/jigsaw-toxic-comment-classification-challenge}{Jigsaw Dataset: Toxic Comment Classification}}, Reddit, and Twitter datasets. For every toxic sentence, a paraphrase, i.e., a non-toxic sentence, is provided.
    The dataset includes two sets, one for training with 11,939 sentence pairs and the other for testing with 671 pairs.  We use the test split for 
    evaluation.
    \item \textbf{APPDIA}\footnote{\href{https://github.com/sabithsn/APPDIA-Discourse-Style-Transfer/tree/main/data/original-annotated-data}{APPDIA Dataset}} is a parallel corpus of close to 2,000 offensive Reddit comments and their paraphrased counterparts annotated by expert sociolinguists. A sentence is offensive if it consists of insults, profane words, hate speech, or threats of violence. The dataset is split into 80-10-10 sets for training, validation, and testing, respectively. We consider the test set for our experiments.
\end{itemize}

\noindent \textbf{Evaluation Metrics.} 
We perform 
reference-free evaluation which is used in style transfer works \cite{logacheva2022paradetox, floto2023diffudetox}.

\begin{itemize}[noitemsep]
    \item \textbf{Style Accuracy (STA)} measures the percentage of outputs classified as non-toxic by a style classifier that is a RoBERTa model fine-tuned on the Jigsaw dataset by \citet{logacheva2022paradetox}.
    \item \textbf{Content Preservation (SIM)} estimates the similarity between the embeddings of the original text and the output using a Paraphrase model by \citet{wieting2019beyond}. This model is trained on paraphrase pairs from the ParaNMT corpus \cite{wieting2017paranmt}. The training objective is to produce embeddings that exhibit higher similarity for paraphrases compared to non-paraphrased sentences.
    \item \textbf{Fluency (FL)} measures the percentage of grammatically correct sentences that are identified by a RoBERTa-based linguistic acceptability classifier trained on the CoLA dataset by \citet{warstadt2019cola}.
\end{itemize}

Following \citet{logacheva2022paradetox}, we also use a joint metric, i.e., the \textbf{\textit{J}} score, which is the multiplication of the three metrics described above as follows:
\begin{equation}
\label{equ:j-score}
    J = \mbox{STA} \times \mbox{SIM} \times \mbox{FL}
\end{equation}


%

Finally, we overlook word-overlap evaluation metrics, such as \textsc{Bleu} \cite{papineni2002bleu}, based on an in-depth analysis by \cite{liu-etal-2016-evaluate} which argues that such metrics are not appropriate for text generation evaluation. \\

\noindent \textbf{EICL Scoring Function.} For ensemble prompting, we use two scoring functions that are: 

\begin{itemize}[noitemsep]
    \item \textbf{S-score:} The style transformation introduces a trade-off between STA and SIM. That is, an output can be classified as non-toxic while its SIM score is lower due to the changes made to the sentence. Conversely, if the output remains very similar to the input, it might get classified as toxic. Therefore, we define the \textit{S-score} as the average of STA and SIM to capture that trade-off:
    \begin{equation}
        S = average(\mbox{STA}, \mbox{SIM})
    \end{equation}
    \item \textbf{J-score:} It incorporates all performance criteria for a comprehensive evaluation, as shown in Eq.\ (\ref{equ:j-score}).
\end{itemize}

Since all metrics are reference-free, our EICL approach can be used for any real-time inference.





\subsection{Experimental Results}

\noindent N.B. Due to the page limit, we only report the results for the best $k$ (denoted by $k^*$) for our in-context learning settings.

\subsubsection{Zero-Shot and Few-Shot Prompting}
\label{sec:zero_and_few_experiments}



\noindent \textbf{ParaDetox.} Table \ref{tab:paradetox-results} compares our \textsc{GPT-Detox} variants with the unsupervised and supervised models from \cite{logacheva2022paradetox, floto2023diffudetox}.
Our few-shot solutions achieve the best performance with respect to the joint metric $J$, also they show promising results for the individual metrics when compared with the baselines.
The CMES solution results in 9\% and 10\% absolute improvement in $J$ when compared with ParaDetox and DiffuDetox, respectively.
Additionally, our solutions have the highest fluency scores being at least 7\% greater than the second-best results.

The few-shot solutions present comparable, but slightly lower, style accuracy (STA) and fluency (FL) values against the zero-shot solution. Despite that, they lead to greater $J$ values since few-shot prompting improved content preservation (SIM) by more than 11\%. The zero-shot solution achieves the highest STA and FL results among all \textsc{GPT-Detox} settings while falling behind for SIM. 
We provide in-depth qualitative analysis and discussion in Section \ref{sec:qualitative_analysis}.

\definecolor{Gray}{gray}{0.9}
\begin{table}[t!]
    \small 
    \centering
    \resizebox{.85\linewidth}{!}{
        \begin{tabular}{l||c|c|c||c}
            \hline
            \rowcolor{gray!30} \multicolumn{5}{c}{\textbf{ParaDetox}} \\ \hline
             & \textbf{STA} & \textbf{SIM} & \textbf{FL} & \textbf{J} \\ \hline
            \textbf{Human} & 0.96 & 0.77 & 0.88 & 0.66\\ \hline 
            \multicolumn{5}{c}{\textbf{Unsupervised}} \\ \hline
            DRG-Template & 0.90 & 0.82 & 0.69 & 0.51\\
            DRG-Retrieve & 0.97 & 0.36 & 0.86 & 0.31\\ 
            Mask\&Infill & 0.91 & 0.82 & 0.63 & 0.48\\
            CondBERT & 0.98 & 0.77 & 0.82 & 0.62\\
            SST & 0.86 & 0.57 & 0.19 & 0.10\\
            ParaGedi & \textbf{0.99} & 0.71 & 0.88 & 0.62\\
            DLSM & 0.76 & 0.76 & 0.52 & 0.25 \\ \hline
            \multicolumn{5}{c}{\textbf{Supervised}} \\ \hline
            ParaDetox & 0.89 & 0.86 & 0.89 & 0.68\\
            DiffuDetox & 0.92 & \textbf{0.88} & 0.80 & 0.67\\ \hline \hline
            \rowcolor{gray!10} \multicolumn{5}{c}{\textbf{\textsc{GPT-Detox} (Prompt-based via GPT-3.5)}} \\ \hline
            {Zero-Shot} & 0.97 & 0.74 & \textbf{0.99} & 0.70 \\ \hline
            \multicolumn{1}{l}{Few-Shot} & \multicolumn{3}{c}{} \\ \hline
            {Random ($k^*=5$)}& 0.92 & 0.85  & 0.97 & 0.75  \\ \hline            
            {WMES ($k^*=10$)} & 0.90 & \textbf{0.88} & 0.96 & \textbf{0.77}\\ \hline            
            {CMES ($k^*=10$)} & 0.93 & 0.87 & 0.96 & \textbf{0.77} \\ \hline
        \end{tabular}%
    }
    \caption{Text detoxification results on the ParaDetox dataset. The unsupervised and supervised baselines are from \cite{logacheva2022paradetox}. $k^*$ represent the best $k$ for the corresponding setting.}
    \label{tab:paradetox-results}
\end{table}

\vspace{6pt}

\noindent \textbf{APPDIA.}
We present the results of our experiments for the APPDIA dataset in Table \ref{tab:appdia-results}\footnote{For the APPDIA experiment, we use the outputs from the baselines in \cite{atwell2022appdia} to obtain the results for all metrics.}. Our \textsc{GPT-Detox} solutions outperform BART and DialoGPT considering the STA, FL, and $J$ with a significant margin. Our zero-shot setting got the highest STA and FL scores. The CMES shows comparable $J$ results when compared to T5. Similar to our earlier observation for the ParaDetox dataset, few-shot prompting outperforms zero-shot prompting because of greater SIM values.  
Our qualitative analysis in Section \ref{sec:qualitative_analysis} provides extensive insights.

\begin{table}[t!]
    \small 
    \centering
    \resizebox{0.85\linewidth}{!}{
        \begin{tabular}{l||c|c|c||c}
            \hline
            \rowcolor{gray!30} \multicolumn{5}{c}{\textbf{APPDIA}} \\ \hline
             & \textbf{STA} & \textbf{SIM} & \textbf{FL} & \textbf{J} \\ \hline
             Human &0.87   & 0.77  &0.95    & 0.65\\ \hline
            \multicolumn{5}{c}{\textbf{APPDIA Baseline Models}} \\ \hline
            BART & 0.73 & \textbf{0.88} & 0.96& 0.62\\
            T5 & 0.82& \textbf{0.88} & \textbf{0.98} & \textbf{0.70}\\
            DialoGPT & 0.86& 0.70& 0.82& 0.47\\ \hline \hline
            \rowcolor{gray!10} \multicolumn{5}{c}{\textbf{\textsc{GPT-Detox} (Prompt-based via GPT-3.5)}} \\ \hline
            {Zero-shot} & \textbf{0.97} & 0.64 & \textbf{0.98} & 0.61\\ \hline
            \multicolumn{1}{l}{Few-Shot} & \multicolumn{3}{c}{} \\ \hline
            {Random ($k^*=8$)} & 0.92 & 0.79 & 0.96 & 0.69\\ \hline
            {WMES ($k^*=10$)} & 0.86 & 0.83 & \textbf{0.98} & \textbf{0.70}\\ \hline
            {CMES ($k^*=10$)} &0.93 & 0.81 & 0.93 & \textbf{0.70}\\ \hline
        \end{tabular}%
    }
    \caption{Text detoxification results on the APPDIA dataset. The baselines are from \cite{atwell2022appdia}. $k^*$ represent the best $k$.}
    \label{tab:appdia-results}
\end{table}

\subsubsection{Ensemble In-context Learning}
\label{sec:experiments-ensemble}

The previous section shows that our base prompting settings, i.e., zero-shot and few-shot, outperformed the state-of-the-art models, with respect to the individual and joint metrics, on the ParaDetox and APPDIA datasets. We now study whether we can improve the results more by leveraging ensemble in-context learning (EICL).
Table \ref{tab:ensemble-results} reports the results. 
%

As for ParaDetox, the EICL variants yield the greatest STA, SIM, and FL. The ensemble scoring functions $S$-score and $J$-score lead to 9\% and 11\% improvement for the overall joint evaluation metric $J$, respectively, compared to the best base prompt settings.
A similar observation is held for the APPDIA dataset regarding the individual metrics. The scoring function $J$-score leads to a 17\% improvement, and $S$-score boosts the performance by 12\%.

Since the scoring function $S$-score considers only STA and SIM, it helps in achieving the greatest value for SIM; whereas, the scoring function $J$-score results in the highest FL because it considers all three individual metrics including FL.
It is worth noting that we maxed out the STA and FL metrics for Paradetox and APPDIA by ensembling.
Finally, this experiment shows that an ensemble of multiple prompts is necessary to capture different aspects of detoxification. 




\definecolor{Gray}{gray}{0.9}
\begin{table}[t!]
    \small 
    \centering
    \resizebox{0.85\linewidth}{!}{
        \begin{tabular}{l||c|c|c||c}
            \hline
            \multicolumn{5}{c}{\textbf{\textsc{GPT-Detox} (ICL vs.\ EICL)}} \\ \hline
            & \textbf{STA} & \textbf{SIM} & \textbf{FL} & \textbf{J} \\ \hline
            \rowcolor{gray!30} \multicolumn{5}{c} {\textbf{ParaDetox}} \\ \hline
            {Zero-Shot} & 0.97 & 0.74 & 0.99 & 0.70 \\ \hline            
            {CMES ($k^*=10$)} & 0.93 & 0.87 & 0.96 & 0.77 \\ \hline
            \rowcolor{gray!10} \multicolumn{5}{l}{{EICL (Ensemble)}} \\ \hline
            {$f_s$ = $S$-score} & \textbf{0.99} & \textbf{0.91} & 0.95 & 0.86\\ \hline
            {$f_s$ = $J$-score} & \textbf{0.99} & 0.90 & \textbf{1.00} & \textbf{0.89}\\ \hline \hline
            \rowcolor{gray!30} \multicolumn{5}{c}{\textbf{APPDIA}} \\ \hline 
            {Zero-shot} & 0.97 & 0.64 & 0.98 & 0.61\\ \hline
            {CMES ($k^*=10$)} & 0.93 & 0.81 & 0.93 & 0.70\\ \hline
            \rowcolor{gray!10} \multicolumn{5}{l}{{EICL (Ensemble)}} \\ \hline
            {$f_s$ = $S$-score} &\textbf{1.00} & \textbf{0.88} & 0.94 & 0.82\\ \hline
            {$f_s$ = $J$-score} & \textbf{1.00} & 0.87 & \textbf{1.00} & \textbf{0.87}\\ \hline
        \end{tabular}%
    }
    \caption{Text detoxification results for in-context learning (ICL) vs.\ ensemble in-context learning (EICL) on Paradetox and APPDIA datasets. $k^*$ represent the best $k$.}
    \label{tab:ensemble-results}
\end{table}

\subsubsection{Human Evaluation}

Measuring the quality of text detoxification is subjective because it involves balancing style accuracy, content similarity, and fluency. 
To complement our automatic evaluation presented in Sections 
\ref{sec:zero_and_few_experiments} and \ref{sec:experiments-ensemble}, we also conduct human evaluation by hiring three annotators being familiar with the task.
For that purpose, the annotators are shown the toxic input, for every test example, and are asked to choose the best non-toxic paraphrase from an anonymized pair of outputs. 
The outputs are from ParaDetox for the ParaDetox dataset, T5 for APPDIA,
and our best-performing setting, i.e., CMES ($k$ = 10).
The labelling exercise is done for three metrics: style accuracy (STA), content preservation (SIM), and fluency (FL) previously defined in Section \ref{sec:exp_settings}.

Table \ref{tab:human_eval} shows that the annotators prefer the responses that are generated by our \textsc{GPT-Detox} CMES ($k = 10$) setting for over 75\% of times across all the metrics for both datasets.
Additionally, based on Cohen's Kappa coefficient \cite{mchugh2012interrater}, the annotators demonstrate moderate to substantial agreement with values ranging from $0.45$ to $0.76$, across all metrics.

\newcolumntype{P}[1]{>{\centering\arraybackslash}p{#1}}
\begin{table}[ht]
    \centering
    \fontsize{9}{11}\selectfont
    \begin{tabular}{P{1.25cm}|P{1.25cm}|P{1.25cm}}
        \hline
        \textbf{STA} & \textbf{SIM} & \textbf{FL} \\ \hline
        \rowcolor{gray!30} \multicolumn{3}{c}{\textbf{ParaDetox}} \\
        \rowcolor{gray!30} \multicolumn{3}{c}{\textbf{CMES ($k=10$) $>$ ParaDetox}} \\ \hline
        0.76 & 0.81 & 0.84 \\ \hline \hline
        \rowcolor{gray!30} \multicolumn{3}{c}{\textbf{APPDIA}} \\
        \rowcolor{gray!30} \multicolumn{3}{c}{\textbf{CMES ($k=10$) $>$ T5}} \\ \hline
        0.78 & 0.80 & 0.76 \\ \hline
    \end{tabular}
    \caption{Human Evaluation}
    \label{tab:human_eval}
\end{table}

Finally, this experiment confirms that LLMs, such as GPT-3.5 Turbo, that are instruction-tuned are not only capable of text detoxification but also they generate responses that are well-aligned with human expectations.

\begin{table*}[t]
    \setlength\dashlinedash{1pt}
    \centering
    \resizebox{\linewidth}{!}{
        \begin{tabular}{l|p{14cm}|c|c}
        \hline
        \rowcolor{gray!30} \multicolumn{2}{c|}{\textbf{ParaDetox}} & S & J \\ \hline
        Input & \textcolor{red}{dont you ever change one of my edits, or i will come to your house , and i will cut you.} & - & -  \\ \cdashline{1-4}
        Zero-shot & Please refrain from modifying any of my edits, as I kindly ask for your cooperation. If changes are made, it may result in a negative consequence. & 0.75 & 0.50 \\
        Random & don't you ever change one of my edits, or I will come to your house, and \textcolor{red}{\hl{I will harm you.}} & 0.48 & 0.0  \\
        \rowcolor{teal!7.5} WMES* & \textcolor{teal}{don't you ever change one of my edits, or i will come to your house, and i will confront you.} & 0.98 & 0.95 \\
        CMES & don't ever change one of my edits, or there will be consequences. & 0.85 & 0.69 \\ \hline
        Input & \textcolor{red}{not even from brainless, hypocritical marxist pigs.} & - & -  \\ \cdashline{1-4}
        \rowcolor{teal!7.5} Zero-shot* & \textcolor{teal}{not even from individuals who lack critical thinking and hold contradictory views influenced by marxist ideology.} & 0.80 & 0.60  \\
        Random & not even from individuals with differing political beliefs. & 0.75 & 0.49 \\
        WMES & not even from \textcolor{red}{\hl{brainless, hypocritical supporters}} of Marxist ideology. & 0.43 & 0.0 \\
        CMES & not even from \textcolor{red}{\hl{hypocritical people}} who follow Marxist ideology. & 0.37 & 0.0  \\ \hline \hline
        \rowcolor{gray!30} \multicolumn{2}{c|}{\textbf{APPDIA}} & S & J \\ \hline
        Input & \textcolor{red}{fuck off dude. worst of the fucking worst!} & - & - \\ \cdashline{1-4}
        Zero-shot & Please leave, my friend. \textcolor{red}{\hl{You are the least desirable of all.}} & 0.21 & 0.0 \\
        Random & please leave dude. one of the worst! & 0.89 & 0.78 \\
        WMES & please leave, dude. \textcolor{red}{\hl{you are one of the worst!}} & 0.35 & 0.0 \\
        \rowcolor{teal!7.5} CMES* & \textcolor{teal}{please leave dude. one of the worst!} & 0.89 & 0.78 \\ \hline
        Input & \textcolor{red}{i don ’ t give them the respect of being called women. they deluded and demented cunts.} & - & - \\ \cdashline{1-4}
        \rowcolor{teal!7.5} Zero-shot* & \textcolor{teal}{i don't acknowledge their identity as women with respect. they have misguided and troubled perspectives.} & 0.87 & 0.74 \\
        Random & i don't give them the respect of being called women. they are \textcolor{red}{\hl{deluded}} and misguided individuals. & 0.48 & 0.0 \\
        WMES & i don't give them the respect of being called women. they are \textcolor{red}{\hl{deluded}} and misguided individuals. & 0.48 & 0.0 \\
        CMES & i don't give them the respect of being called women. they are \textcolor{red}{\hl{deluded}} and misguided individuals. & 0.48 & 0.0  \\ \hline
        \end{tabular}
    }
\caption{Examples for Qualitative Analysis. The green highlight shows the best prompt. The yellow highlight shows toxic parts in some generated outputs.}
\label{tab:qual_analysis}
\end{table*}

\begin{table*}[h!]
    \setlength\dashlinedash{1pt}
    \centering
    \resizebox{\linewidth}{!}{
    \begin{tabular}{l|p{14.5cm}}
    \hline
    \rowcolor{gray!30} \multicolumn{2}{c}{\textbf{ParaDetox}} \\
    \hline
    Input & \textcolor{red}{west is the spoiled shit.} \\
    Reference & west is bad \\
     \hline
     ParaDetox & west is spoiled. \\
     \cdashline{1-2}
    Zero-shot & West is a person who has been given many privileges and may have a tendency to act entitled. \\
    Random & west is the spoiled person.\\
    WMES & west is the spoiled one. \\
    CMES & The west is spoiled. \\    
    \hline
    \rowcolor{gray!30} \multicolumn{2}{c}{\textbf{APPDIA}} \\ \hline
    Input &  \textcolor{red}{i wish i could spit my unvaccinated saliva on you and your family.} \\
    Reference & i wish i could weaponize my unvaccinated saliva. \\
     \hline
     T5 & i wish i could be more unvaccinated . \\
     \cdashline{1-2}
    Zero-shot & I hope that my saliva, which has not been vaccinated, does not come into contact with you or your loved ones. \\
    Random & I wish you and your family well.\\
    WMES & I wish you and your family well. \\
    CMES & i wish you and your family could be safe from my unvaccinated saliva . \\
    \hline
    \end{tabular}
    }
    \caption{Failed cases by the baselines and \textsc{GPT-Detox}}
    \label{tab:bad_examples}
\end{table*}

\subsubsection{Qualitative Analysis}
\label{sec:qualitative_analysis}
This section presents qualitative examples for both of the datasets in Table \ref{tab:qual_analysis}.
Under all the settings, zero-shot prompts produce verbose text that is a complete paraphrase of the original sentence. 
Those responses are fluent and often non-toxic.
However, the pre-trained similarity model gives zero-shot outputs lower SIM scores due to drastic style changes. 
On the other hand, the presence of examples guides the few-shot prompts to generate non-toxic paraphrases with few changes in the style. 
This leads to an increase in the SIM metric and consequently an increase of $S$-score and $J$-score used for ensembling as well.
%
However, our few-shot settings tend to retain toxicity as a side-effect of stronger content preservation which results in the $J$ value being 0 more often compared to the zero-shot setting.

Table \ref{tab:bad_examples} shows examples where the \textsc{GPT-Detox} variants fail to detoxify the input text appropriately.
The input ``\textit{west is the spoiled shit.}'' from the ParaDetox dataset is an ambiguous example where ``\textit{west}'' may refer to either a geographical concept or a famous person (e.g., the rapper Kanye West).
Although the input refers to the geographical concept, our \textsc{GPT-Detox} settings toggle between the aforesaid concepts.
As to the APPDIA example, i.e., ``\textit{i wish i could spit my unvaccinated saliva on you and your family.}'', all our settings negate the intention of the toxic input and generate an incorrect non-toxic response. Also, T5 suffers from this confusion and generates an unrelated paraphrase. These examples indicate that GPT-3.5 Turbo may still have room for improvement.

\section{Conclusion}
\label{sec:conclusion}

We introduced \textsc{GPT-Detox} as a prompt-based in-context learning framework for text detoxification with GPT-3.5 Turbo. We established two example selection techniques, namely, word-matching example selection (WMES) and context-matching example selection (CMES), for few-shot prompting.
We also considered ensemble in-context learning (EICL).
In our extensive experiments, we compared our zero-shot, few-shot, and EICL solutions with state-of-the-art models against the ParaDetox and APPDIA datasets. 
The zero-shot solution achieved the highest STA and FL values, whereas, few-shot settings led to better overall performance by improving SIM.
The ensemble solution obtained the greatest performance across all metrics and demonstrated the necessity of using multiple prompts to capture different aspects of detoxification.
We also provided a comprehensive qualitative analysis that discusses the pros and cons of all of our settings.
%

In future work, we will investigate the performance of GPT-4 for text detoxification. We are also interested in the other applications of in-context learning in NLP. Finally, we like to investigate the potential use of LLMs for AI Evaluation.

{\footnotesize
\bibliography{icmla2023}}

\begin{thebibliography}{34}
\providecommand{\natexlab}[1]{#1}
\providecommand{\url}[1]{#1}
\csname url@samestyle\endcsname
\providecommand{\newblock}{\relax}
\providecommand{\bibinfo}[2]{#2}
\providecommand{\BIBentrySTDinterwordspacing}{\spaceskip=0pt\relax}
\providecommand{\BIBentryALTinterwordstretchfactor}{4}
\providecommand{\BIBentryALTinterwordspacing}{\spaceskip=\fontdimen2\font plus
\BIBentryALTinterwordstretchfactor\fontdimen3\font minus
  \fontdimen4\font\relax}
\providecommand{\BIBforeignlanguage}[2]{{%
\expandafter\ifx\csname l@#1\endcsname\relax
\typeout{** WARNING: IEEEtranN.bst: No hyphenation pattern has been}%
\typeout{** loaded for the language `#1'. Using the pattern for}%
\typeout{** the default language instead.}%
\else
\language=\csname l@#1\endcsname
\fi
#2}}
\providecommand{\BIBdecl}{\relax}
\BIBdecl

\bibitem[Gonz{\'a}lez-Bail{\'o}n and Lelkes(2023)]{gonzalez2023social}
S.~Gonz{\'a}lez-Bail{\'o}n and Y.~Lelkes, ``Do social media undermine social
  cohesion? a critical review,'' \emph{Social Issues and Policy Review},
  vol.~17, no.~1, pp. 155--180, 2023.

\bibitem[Zhao et~al.(2023)Zhao, Zhou, Li, Tang, Wang, Hou, Min, Zhang, Zhang,
  Dong, et~al.]{zhao2023survey}
W.~X. Zhao, K.~Zhou, J.~Li, T.~Tang, X.~Wang, Y.~Hou, Y.~Min, B.~Zhang,
  J.~Zhang, Z.~Dong \emph{et~al.}, ``A survey of large language models,''
  \emph{arXiv preprint arXiv:2303.18223}, 2023.

\bibitem[Weidinger et~al.(2021)Weidinger, Mellor, Rauh, Griffin, Uesato, Huang,
  Cheng, Glaese, Balle, Kasirzadeh, et~al.]{weidinger2021ethical}
L.~Weidinger, J.~Mellor, M.~Rauh, C.~Griffin, J.~Uesato, P.-S. Huang, M.~Cheng,
  M.~Glaese, B.~Balle, A.~Kasirzadeh \emph{et~al.}, ``Ethical and social risks
  of harm from language models,'' \emph{arXiv preprint arXiv:2112.04359}, 2021.

\bibitem[Brown et~al.(2020)Brown, Mann, Ryder, Subbiah, Kaplan, Dhariwal,
  Neelakantan, Shyam, Sastry, Askell, et~al.]{brown2020language}
T.~Brown, B.~Mann, N.~Ryder, M.~Subbiah, J.~D. Kaplan, P.~Dhariwal,
  A.~Neelakantan, P.~Shyam, G.~Sastry, A.~Askell \emph{et~al.}, ``Language
  models are few-shot learners,'' \emph{Advances in neural information
  processing systems}, vol.~33, pp. 1877--1901, 2020.

\bibitem[Ouyang et~al.(2022)Ouyang, Wu, Jiang, Almeida, Wainwright, Mishkin,
  Zhang, Agarwal, Slama, Ray, et~al.]{ouyang2022training}
L.~Ouyang, J.~Wu, X.~Jiang, D.~Almeida, C.~Wainwright, P.~Mishkin, C.~Zhang,
  S.~Agarwal, K.~Slama, A.~Ray \emph{et~al.}, ``Training language models to
  follow instructions with human feedback,'' \emph{Advances in Neural
  Information Processing Systems}, vol.~35, pp. 27\,730--27\,744, 2022.

\bibitem[Christiano et~al.(2017)Christiano, Leike, Brown, Martic, Legg, and
  Amodei]{christiano2017deep}
P.~F. Christiano, J.~Leike, T.~Brown, M.~Martic, S.~Legg, and D.~Amodei, ``Deep
  reinforcement learning from human preferences,'' \emph{Advances in neural
  information processing systems}, vol.~30, 2017.

\bibitem[Li et~al.(2018)Li, Jia, He, and Liang]{li2018delete}
J.~Li, R.~Jia, H.~He, and P.~Liang, ``Delete, retrieve, generate: a simple
  approach to sentiment and style transfer,'' \emph{arXiv preprint
  arXiv:1804.06437}, 2018.

\bibitem[Wu et~al.(2019)Wu, Zhang, Zang, Han, and Hu]{wu2019mask}
X.~Wu, T.~Zhang, L.~Zang, J.~Han, and S.~Hu, ````mask and infill'': Applying
  masked language model to sentiment transfer,'' \emph{arXiv preprint
  arXiv:1908.08039}, 2019.

\bibitem[Lee(2020)]{lee2020stable}
J.~Lee, ``Stable style transformer: Delete and generate approach with
  encoder-decoder for text style transfer,'' \emph{arXiv preprint
  arXiv:2005.12086}, 2020.

\bibitem[Dale et~al.(2021)Dale, Voronov, Dementieva, Logacheva, Kozlova,
  Semenov, and Panchenko]{dale2021text}
D.~Dale, A.~Voronov, D.~Dementieva, V.~Logacheva, O.~Kozlova, N.~Semenov, and
  A.~Panchenko, ``Text detoxification using large pre-trained neural models,''
  \emph{arXiv preprint arXiv:2109.08914}, 2021.

\bibitem[Logacheva et~al.(2022)Logacheva, Dementieva, Ustyantsev, Moskovskiy,
  Dale, Krotova, Semenov, and Panchenko]{logacheva2022paradetox}
V.~Logacheva, D.~Dementieva, S.~Ustyantsev, D.~Moskovskiy, D.~Dale, I.~Krotova,
  N.~Semenov, and A.~Panchenko, ``Paradetox: Detoxification with parallel
  data,'' in \emph{Proceedings of the 60th Annual Meeting of the Association
  for Computational Linguistics (Volume 1: Long Papers)}, 2022, pp. 6804--6818.

\bibitem[Atwell et~al.(2022)Atwell, Hassan, and Alikhani]{atwell2022appdia}
K.~Atwell, S.~Hassan, and M.~Alikhani, ``Appdia: A discourse-aware
  transformer-based style transfer model for offensive social media
  conversations,'' \emph{arXiv preprint arXiv:2209.08207}, 2022.

\bibitem[Liu et~al.(2022{\natexlab{a}})Liu, Gao, Jia, Xu, and
  Vosoughi]{liu2022nonparallel}
R.~Liu, C.~Gao, C.~Jia, G.~Xu, and S.~Vosoughi, ``Non-parallel text style
  transfer with self-parallel supervision,'' 2022.

\bibitem[He et~al.(2020)He, Wang, Neubig, and
  Berg-Kirkpatrick]{he2020probabilistic}
J.~He, X.~Wang, G.~Neubig, and T.~Berg-Kirkpatrick, ``A probabilistic
  formulation of unsupervised text style transfer,'' \emph{arXiv preprint
  arXiv:2002.03912}, 2020.

\bibitem[Lewis et~al.(2020)Lewis, Liu, Goyal, Ghazvininejad, Mohamed, Levy,
  Stoyanov, and Zettlemoyer]{lewis2020bart}
M.~Lewis, Y.~Liu, N.~Goyal, M.~Ghazvininejad, A.~Mohamed, O.~Levy, V.~Stoyanov,
  and L.~Zettlemoyer, ``{BART}: Denoising sequence-to-sequence pre-training for
  natural language generation, translation, and comprehension,'' in
  \emph{Proceedings of the 58th Annual Meeting of the Association for
  Computational Linguistics}.\hskip 1em plus 0.5em minus 0.4em\relax Online:
  Association for Computational Linguistics, Jul. 2020, pp. 7871--7880.

\bibitem[Floto et~al.(2023)Floto, Pour, Farinneya, Tang, Pesaranghader,
  Bharadwaj, and Sanner]{floto2023diffudetox}
G.~Floto, M.~M.~A. Pour, P.~Farinneya, Z.~Tang, A.~Pesaranghader, M.~Bharadwaj,
  and S.~Sanner, ``Diffudetox: A mixed diffusion model for text
  detoxification,'' 2023.

\bibitem[Schick and Schütze(2021)]{schick2021its}
T.~Schick and H.~Schütze, ``It's not just size that matters: Small language
  models are also few-shot learners,'' 2021.

\bibitem[Liu et~al.(2022{\natexlab{b}})Liu, Shen, Zhang, Dolan, Carin, and
  Chen]{liu-etal-2022-makes}
J.~Liu, D.~Shen, Y.~Zhang, B.~Dolan, L.~Carin, and W.~Chen, ``What makes good
  in-context examples for {GPT}-3?'' in \emph{Proceedings of Deep Learning
  Inside Out (DeeLIO 2022): The 3rd Workshop on Knowledge Extraction and
  Integration for Deep Learning Architectures}.\hskip 1em plus 0.5em minus
  0.4em\relax Dublin, Ireland and Online: Association for Computational
  Linguistics, May 2022, pp. 100--114.

\bibitem[Wei et~al.(2021)Wei, Bosma, Zhao, Guu, Yu, Lester, Du, Dai, and
  Le]{wei2021finetuned}
J.~Wei, M.~Bosma, V.~Y. Zhao, K.~Guu, A.~W. Yu, B.~Lester, N.~Du, A.~M. Dai,
  and Q.~V. Le, ``Finetuned language models are zero-shot learners,''
  \emph{arXiv preprint arXiv:2109.01652}, 2021.

\bibitem[Meng et~al.(2022)Meng, Huang, Zhang, and Han]{meng2022generating}
Y.~Meng, J.~Huang, Y.~Zhang, and J.~Han, ``Generating training data with
  language models: Towards zero-shot language understanding,'' 2022.

\bibitem[Yoo et~al.(2021)Yoo, Park, Kang, Lee, and Park]{yoo2021gpt3mix}
K.~M. Yoo, D.~Park, J.~Kang, S.-W. Lee, and W.~Park, ``Gpt3mix: Leveraging
  large-scale language models for text augmentation,'' \emph{arXiv preprint
  arXiv:2104.08826}, 2021.

\bibitem[Zheng and Huang(2022)]{zheng2022exploring}
C.~Zheng and M.~Huang, ``Exploring prompt-based few-shot learning for grounded
  dialog generation,'' 2022.

\bibitem[Madotto et~al.(2021)Madotto, Lin, Winata, and
  Fung]{madotto2021fewshot}
A.~Madotto, Z.~Lin, G.~I. Winata, and P.~Fung, ``Few-shot bot: Prompt-based
  learning for dialogue systems,'' 2021.

\bibitem[Ye et~al.()Ye, Peters, Ren, and Hajishirzi]{yeinvestigating}
Q.~Ye, I.~B. M.~E. Peters, X.~Ren, and H.~Hajishirzi, ``Investigating fusion
  methods for in-context learning.''

\bibitem[Lee et~al.(2022)Lee, Lim, and Choi]{lee2022does}
Y.-J. Lee, C.-G. Lim, and H.-J. Choi, ``Does gpt-3 generate empathetic
  dialogues? a novel in-context example selection method and automatic
  evaluation metric for empathetic dialogue generation,'' in \emph{Proceedings
  of the 29th International Conference on Computational Linguistics}, 2022, pp.
  669--683.

\bibitem[Ye et~al.(2023)Ye, Hwang, Yang, Yun, Kim, and Seo]{ye2023context}
S.~Ye, H.~Hwang, S.~Yang, H.~Yun, Y.~Kim, and M.~Seo, ``In-context instruction
  learning,'' \emph{arXiv preprint arXiv:2302.14691}, 2023.

\bibitem[Shen et~al.(2023)Shen, Hou, Zhou, Du, Longpre, Wei, Chung, Zoph,
  Fedus, Chen, et~al.]{shen2023flan}
S.~Shen, L.~Hou, Y.~Zhou, N.~Du, S.~Longpre, J.~Wei, H.~W. Chung, B.~Zoph,
  W.~Fedus, X.~Chen \emph{et~al.}, ``Flan-moe: Scaling instruction-finetuned
  language models with sparse mixture of experts,'' \emph{arXiv preprint
  arXiv:2305.14705}, 2023.

\bibitem[Wang et~al.(2023)Wang, Xu, Lan, Hu, Lan, Lee, and Lim]{wang2023plan}
L.~Wang, W.~Xu, Y.~Lan, Z.~Hu, Y.~Lan, R.~K.-W. Lee, and E.-P. Lim,
  ``Plan-and-solve prompting: Improving zero-shot chain-of-thought reasoning by
  large language models,'' 2023.

\bibitem[Wieting et~al.(2019)Wieting, Berg-Kirkpatrick, Gimpel, and
  Neubig]{wieting2019beyond}
J.~Wieting, T.~Berg-Kirkpatrick, K.~Gimpel, and G.~Neubig, ``Beyond bleu:
  training neural machine translation with semantic similarity,'' \emph{arXiv
  preprint arXiv:1909.06694}, 2019.

\bibitem[Wieting and Gimpel(2017)]{wieting2017paranmt}
J.~Wieting and K.~Gimpel, ``Paranmt-50m: Pushing the limits of paraphrastic
  sentence embeddings with millions of machine translations,'' \emph{arXiv
  preprint arXiv:1711.05732}, 2017.

\bibitem[Warstadt et~al.(2019)Warstadt, Singh, and Bowman]{warstadt2019cola}
A.~Warstadt, A.~Singh, and S.~R. Bowman, ``Cola: The corpus of linguistic
  acceptability (with added annotations),'' 2019.

\bibitem[Papineni et~al.(2002)Papineni, Roukos, Ward, and
  Zhu]{papineni2002bleu}
K.~Papineni, S.~Roukos, T.~Ward, and W.-J. Zhu, ``Bleu: a method for automatic
  evaluation of machine translation,'' in \emph{Proceedings of the 40th annual
  meeting of the Association for Computational Linguistics}, 2002, pp.
  311--318.

\bibitem[Liu et~al.(2016)Liu, Lowe, Serban, Noseworthy, Charlin, and
  Pineau]{liu-etal-2016-evaluate}
C.-W. Liu, R.~Lowe, I.~Serban, M.~Noseworthy, L.~Charlin, and J.~Pineau, ``How
  {NOT} to evaluate your dialogue system: An empirical study of unsupervised
  evaluation metrics for dialogue response generation,'' in \emph{Proceedings
  of the 2016 Conference on Empirical Methods in Natural Language
  Processing}.\hskip 1em plus 0.5em minus 0.4em\relax Austin, Texas:
  Association for Computational Linguistics, Nov. 2016, pp. 2122--2132.

\bibitem[McHugh(2012)]{mchugh2012interrater}
M.~L. McHugh, ``Interrater reliability: the kappa statistic,'' \emph{Biochemia
  medica}, vol.~22, no.~3, pp. 276--282, 2012.

\end{thebibliography}


\bibliographystyle{IEEEtranN}

\end{document}